\documentclass{article}
\usepackage{ijcai22}

\usepackage{times}
\usepackage{soul}
\usepackage{url}
\usepackage[hidelinks]{hyperref}
\usepackage[utf8]{inputenc}
\usepackage[small]{caption}
\usepackage{graphicx}
\usepackage{amsmath}
\usepackage{amsthm}
\usepackage{booktabs}
\usepackage{algorithm}
\usepackage{algorithmic}
\usepackage{CJKutf8}
\usepackage{multirow}
\usepackage{makecell}
\usepackage{listings}
\title{Prompt-based Generative Approach towards Multi-Hierarchical Medical Dialogue State Tracking}

\author{
Jun Liu $^1$\and
Tong Ruan $^1$\and
Haofen Wang $^2$\And
Huanhuan Zhang $^1$
\affiliations
$^1$ East China University of Science and Technology\\
$^2$ Tongji University\\
\emails
Y30201043@mail.ecust.edu.cn,
ruantong@ecust.edu.cn,
carter.whfcarter@gmail.com,
hzhang@ecust.edu.cn
}
\begin{document}

\maketitle

\begin{abstract}

%%抽象方式描述 出现的问题 （术语来自不同的句子。。

%%抽什么，更复杂。。。术语不连贯，来自不同的句子

%%状态比较丰富

%%我们在临床场景下，生成了数据集，用了xxx方法，

%%注重generative based模型，什么什么好处

The medical dialogue system is a promising application that can provide great convenience for patients. The dialogue state tracking (DST) module in the medical dialogue system which interprets utterances into the machine-readable structure for downstream tasks is particularly challenging. Firstly, the states need to be able to represent compound entities such as symptoms with their body part or diseases with degrees of severity to provide enough information for decision support. Secondly, these named entities in the utterance might be discontinuous and scattered across sentences and speakers. These also make it difficult to annotate a large corpus which is essential for most methods. Therefore, we first define a multi-hierarchical state structure. We annotate and publish a medical dialogue dataset in Chinese. To the best of our knowledge, there are no publicly available ones before. Then we propose a Prompt-based Generative Approach which can generate slot values with multi-hierarchies incrementally using a top-down approach. A dialogue style prompt is also supplemented to utilize the large unlabeled dialogue corpus to alleviate the data scarcity problem. The experiments show that our approach outperforms other DST methods and is rather effective in the scenario with little data.

\end{abstract}

\section{Introduction}

% \begin{figure*}[htbp] %H为当前位置，!htb为忽略美学标准，htbp为浮动图形
% \centering %图片居中
% \includegraphics[width=1\textwidth]{overview.png} %插入图片，[]中设置图片大小，{}中是图片文件名
% \caption{Workflow of our approach. The input is a dialogue and output is the dialogue state. We treat the process of extracting multi-hierarchies state as a multi-turn QA and compose all the answers as the final output.} %最终文档中希望显示的图片标题
% \label{Fig.main1} %用于文内引用的标签
% \end{figure*}

\begin{table*}
\centering
\begin{tabular}{lll}
\hline \textbf{Input Dialogue} $D$ & \textbf{State Definition} & \textbf{Output State}  \\ \hline 
\makecell[l]{(ours) \\
$U_1$  \textbf{Patient}:  I feel uncomfortable. \\
$R_1$ \textbf{Doctor}:  What's wrong with you? \\
$U_2$ \textbf{Patient}:  My \textit{head} sometimes \textit{feels painful}. \\
% \quad \quad \quad \quad \; \;  What can I do?\\
} &
\makecell[l]{
$I$: \{\textit{describe}, \textit{ask}, \textit{response}, ... \}\\
$S$: \{ \\
\quad symptom: \{ \textit{name}, \textit{extent}, \textit{if\_exists} \} , \\
\quad disease: \{ \textit{name}, \textit{if\_exists} \} \\
\quad ... \\
\} 
% $F($describe$)$ = \{symptom, disease, ...\} \\
% $F($ask$)$ = \{...\}
} &
\makecell[l]{ 
 "intent": ["describe"], \\
 "slot": \{ \\
\quad     "symptom": [\{ \\ 
\quad \quad  "name": "head feels painful", \\ 
\quad \quad  "extent": "sometimes" \\
\quad \quad  "if\_exists": \textit{true}, \\
\quad      \}], \\ 
\quad     "disease": [] \\ 
  \} 
} \\
\hline 
\makecell[l]{
(common DST task) \\
$U_1$  \textbf{User}:  I'd like to book a flight to LA. \\
$R_1$ \textbf{System}:  What date do you want to depart? \\
$U_2$ \textbf{User}:  June 18th. \\
} &
\makecell[l]{
$I$: \{\textit{book\_flight}, \textit{find\_flight}\}\\
$S$: \{ \\
\quad day, \\
\quad departure, \\
\quad destination, \\
\quad ... \\
\}
} &
\makecell[l]{ 
 "intent": "book\_flight", \\
 "slot": \{ \\
\quad     "day": "June 18th", \\
\quad     "departure": "", \\
\quad     "destination": "LA" \\ 
  \} 
} \\
\hline
\end{tabular}
\caption{\label{state-sample} Samples of input dialogue, state definition, and output. The task is to convert the left input dialogue into the right structured state according to the middle definition. The structure of "symptom" in the output state is the same with $S[symptom]$ in the definition. The one in the first row is the situation we are facing and the one below is the common DST Task which state is simpler. The main difference is that our state definition is hierarchical. } 
\end{table*}

%%问题 和 挑战，现有解决，新问题特殊性，contribution 123

%% 医疗对话有未来，有什么方法。。mdst是核心问题，讲难点，我提出了模型，做到了什么地步，看看数据集

% 对话系统很重要，需要理解用户语言，包含意图和对象->两个任务，slot预定义 xxxx通过捕获填充，最后形成state，要跟踪，最后该处action
% 医疗 slot 多 value 大 导致空间大所以DST 有挑战
% 我们这份工作用了xxx技术

The medical dialogue system that can simulate the actions of clinicians is a promising application. The systems acquire the patient's information interactively through natural language and give diagnoses and suggestions of treatments. One of the major challenges in implementing such a system is how to understand the patient's utterances and convert the utterances into a machine-friendly structure. In particular, the structure mainly consists of two parts, patience's major intention and the detailed values of the intention, called intent and slot respectively. The task of defining, extracting, and maintaining such a structure is called Dialogue State Tracking (DST) in task-oriented dialogue systems.

Current task-oriented dialogue systems(\cite{lei-etal-2018-sequicity}, \cite{acharya-etal-2021-alexa}) mainly focus on simple personal assistant tasks such as buying movie tickets, booking restaurants. Though logically the research on medical dialogue systems (\cite{Kao_Tang_Chang_2018}, \cite{wei-etal-2018-task}  \cite{Xu_Zhou_Gong_Liang_Tang_Lin_2019}) exist, the tasks are processed in the same way as simpler systems. However, after we began with a project on a dialogue-based medical decision support system one year before, we found the differences and challenges obvious. 

% There are varies of methods for the DST tasks. 
% Extraction and Generation based. 
% learning-based methods need Corpus  /low resources.
% There are three challenges
%确少

%Major advantages take places 
%what the  which includes intent and slot two parts. The intent is used to indicate what the user is doing, like describing or asking, which is important for how the system responds. Slots include the related information about the intent, like what symptoms the user is describing. The slot is predefined and the value is identified from the input to fill the slot. In other words, the DST task includes intent detection and slot filling two sub-tasks. As the conversation progresses, the processed result of each utterance constitutes the dialogue state and the whole task is called Dialogue State Tracking(DST).
% 现在通用系统，做到什么地步。用在medical上的问题，

%However, some of the differences between the medical domain and general domain lead to the fact that the medical dialogue state tracking is not sufficient to support the medical dialogue system.

% 引文 目前的状态
% 现在是怎么解决不了的

% Many task-oriented systems have been a large-scale online services, but medical dialogue system is still difficult to be used widely. The main reason is that medical dialogue system has low tolerance for mistakes which brings more challenges than normal task-oriented systems. 

%%问题 先抽象 cross sentence/role ner extraction 非连续 

%%形成了多层 带来了困难 标注/抽取 low resource
Firstly, it requires more complex and accurate state information to make the correct diagnosis possible. Typical state structure of task-oriented systems is in the form of \textit{\{intent,(slot, value)+\}}. A case in point in the restaurant reservation system may be \textit{\{book, (time, 5pm), (day, Jan.13)\}}. However, simple slot and slot-values pairs are not enough for medical systems. For example, the patient says "I do not have a headache", it's wrong to convert the utterance to \textit{("symptom", "headache")} and it is a disaster to neglect the negation in the medical domain. Furthermore, if another patient says "I have a severe headache after drinking", Both the severity and condition of the symptom should be extracted and represented. Besides, the number of the intention types of the patients and clinicians are much bigger. In real clinical practice, the patients are not only in need of diagnosis or treatment, but also acquire medical knowledge or look for the console. Typical samples for comparison can be found in Table \ref{state-sample}.  

\begin{CJK*}{UTF8}{gbsn}
Secondly, there are many discontinuous, overlapping, and nested entities in the utterances. These entities may even across sentences and speakers. For example, the patient says "我的头感到有点疼 (My head feels painful)". Most named entity recognition (NER) model will extract "疼 (painful)", however, the right answer should be "头疼 (headache)", and "头" "疼" are separated in original sentences. Another example is "伤口又疼又痒 (wound hurts and itches)". Usually NER model will treat this as a whole entity, but to be accurate, we want "伤口疼 (wound hurts)" and "伤口痒 (wound itches)" so that it will be possible to be normalized into two symptoms. 
\end{CJK*}

Traditionally most existing DST methods are based on deep learning and require a large annotated dataset. For the medical dialogue application, both the complexity of state structure and diverse forms of named entities imply the requirement of a larger corpus. However, it is almost unrealistic due to the annotation cost.

To tackle the problems above, we first redefine the state structure with multiple hierarchies. The fine-grained representation can describe more detailed and semantically related information. Then we choose the generative model to extract both the intent and different types of slot values incrementally and uniformly. A dialogue-style prompt is proposed which transforms the DST task into a response generation task. This also makes it possible to pretrain the model on unlabeled dialogue to alleviate the problem of lacking annotated data. A constrained decoding algorithm is also used to avoid the model from generating invalid output. The experiment shows our method greatly outperforms the baseline methods.

% In this study, we solve the DST problem using a generative model. We first construct a question according to the target intents or slots, input the question along with the dialogue to the model and let the model generate the answer which also is the intent or slot value. The questions are made of natural language which makes it easy to represent complicated state structure. The flexibility of output also makes the model able to deal with intent detection and slot filling at the same tiem, which also makes it easy to transfer between different departments. To better solve the task, we pretrain the model to make it adapt to the medical dialogue and put on extra limits during the process of decoding to avoid the model generating invalid result.

Our contribution can be listed as follows:
% 重点强调
\begin{itemize}
\item We start the first substantial step to extend the current task-oriented dialogue system with complex medical decision support. In particular, we define the dialogue state with multiple hierarchies which is much complex compared to the state of personal assistant tasks. We also annotate a dataset based on the definition. To the best of our knowledge, this is the first well-annotated dataset on medical dialogue systems. 
\item We transform the DST task into a response generation task. We use a generative model with a constrained decoding algorithm to deal with the complexity of state structures and slot values. We also propose a dialogue style prompt that can fully utilize the large unlabeled medical dialogue corpus.
\item The experiments show that our method outperforms the baseline with 6.51\% on the slot filling task and pretraining can improve 4.29\% when only 20\% training data is available. %具体数据
\end{itemize}

% contribution....

 \begin{figure*}[htbp] %H为当前位置，!htb为忽略美学标准，htbp为浮动图形
\centering %图片居中
\includegraphics[width=0.8\textwidth]{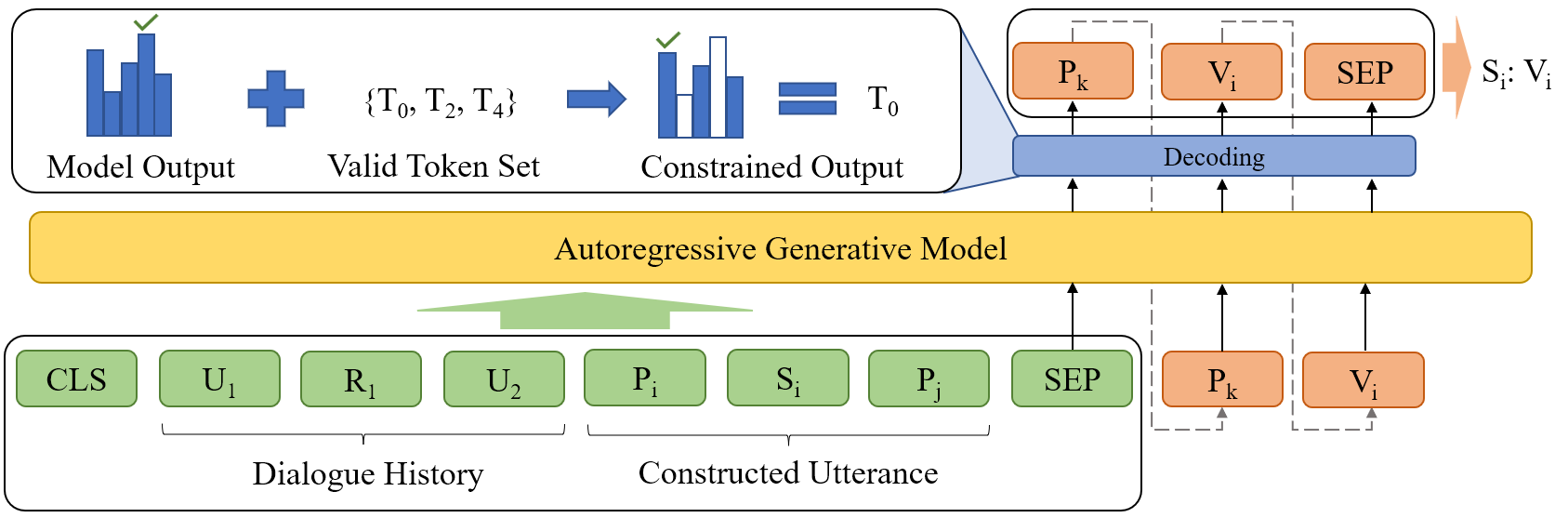} %插入图片，[]中设置图片大小，{}中是图片文件名
\caption{ The illustration of our approach. The input contains two-part, one is the dialogue history and the other is a dialogue-style utterance constructed with prompt token $P_i, P_j, ...$ and target slot type $S_i$. The model will output the response which also contains both prompt token $P_K, ...$ and answer $V_i$. $V_i$ is the value of the input slot $S_i$. A decoding strategy of constructing a valid token set is used to constrain the output. } %最终文档中希望显示的图片标题
\label{arche} %用于文内引用的标签
\end{figure*}

\section{Related Work}

% 方法特点

% 和joint比对
% joint 有什么意义   意图和槽语义上有关联的，所以识别是不同角度，参数共享
% comment 问题
% 新任务 + 现有方法not work + 小改进

% 其他医疗对话系统
% 创新了啥
% DST 为什么对我们要做的不够
% DST 别人的做的不够

% The medical dialogue system that can provide automatic diagnosis is attracting more and more attention due to its promising potentials.
The two subtasks of dialogue state tracking, namely intent detection and slot filling, used to be handled separately. Typically, indent detection is treated as a classification problem and slot filling as a sequence labeling problem(\cite{xu-hu-2018-end}). Recent approaches try to solve them jointly (\cite{Chen_Lv_Wang_Zhu_Tan_Yu_2020}) so that the relationship between the two tasks can be fully exploited. While the two tasks take different forms traditionally, generation-based methods (\cite{WuTradeDST2019}, \cite{kim-etal-2020-efficient}) take the advantage of unifying various types of NLP tasks. \cite{feng-etal-2021-sequence} proposed the Seq2seq-DU method formalizing DST as a sequence to sequence problem using BERT\cite{devlin-etal-2019-bert} and point generation, but its performance may be affected when facing utterances with multiple values. Another reason for the generation-based methods being popular is the publishing of large annotated DST datasets, such as MultiWOZ(\cite{budzianowski2018large}, \cite{ramadan2018large}, \cite{eric2019multiwoz}, \cite{zang2020multiwoz}), Schema-Guided Dialogue Dataset \cite{rastogi2019towards}, etc. A large training dataset is necessary for the generation-based method due to its larger searching spaces.

%Intent detection and slot filling, the two subtasks of dialogue state tracking, used to be handled separately.  treating indent detection as a classification problem and slot filling as a sequence labeling problem(\cite{xu-hu-2018-end}). Recent approaches try to solve them together (\cite{Chen_Lv_Wang_Zhu_Tan_Yu_2020}) so that it will be able to exploit the close relationship between the two tasks. Since these tasks have different forms, using the output flexibility of generation-based methods (\cite{WuTradeDST2019}, \cite{kim-etal-2020-efficient}) is easy to bridge the gaps between them which makes it popular recently. \cite{feng-etal-2021-sequence} proposed Seq2seq-DU formalizing DST as a sequence to sequence problem using BERT\cite{devlin-etal-2019-bert} and point generation, but its performance may be affected when facing utterance with multiple values. Another reason for the generation-based methods being popular is the proposal of large annotated DST datasets, like MultiWOZ(\cite{budzianowski2018large}, \cite{ramadan2018large}, \cite{eric2019multiwoz}, \cite{zang2020multiwoz}), Schema-Guided Dialogue Dataset \cite{rastogi2019towards}, etc. This is necessary for the generation-based method due to its large solving spaces.

But annotating a DST dataset is rather expensive, some studies try to avoid this problem. \cite{lin2021zero} uses multiple machine reading comprehension datasets to train a generative model and achieves zero-shot cross-task transferring to the DST field. \cite{du-etal-2021-qa} uses a weakly supervised method to pretrain a span-based QA model for zero-shot slot filling. But these methods still need large annotated datasets of other tasks which are not always available.

As for the medical field, several dialogue systems (\cite{Kao_Tang_Chang_2018}, \cite{wei-etal-2018-task}, \cite{Xu_Zhou_Gong_Liang_Tang_Lin_2019} ) have been proposed to provide automatic diagnosis.  The setup is similar to the general one except that its contents relate to the medical domain. \cite{DBLP:journals/corr/abs-2010-07497} published a medical consultation dataset on the dialogue system without annotation. \cite{Shi_Hu_Che_Sun_Liu_Huang_2020} created a dataset for the medical slot filling task but only consider one \textit{symptom} slot type with definite slot values.

However, making a diagnosis is more complex than making a decision in the general field like booking a restaurant. In this paper, we fully exploit the difficulties for the DST tasks in the medical dialogue system, providing solutions as well as publishing annotated datasets.   

% DST has two subtasks, intent detection and slot filling. Several datasets are constructed according to that like  These datasets require the model to find what domain the user is currently interested in and extract the related information, like the user is concerned about the flight and the related information might be the departure time.

% Most DST datasets are built for general purpose. In medical field, few dataset is available. 

% One of the shortcomings of generation-based methods is that the model might generate any words which is likely to be invalid. \cite{gao-etal-2020-machine} proposed a canonicalization technique to find the most similar result in valid result set for each model prediction result, but this approach needs a pre-defined result set which is not always possible and cannot utilize model output.  \cite{lee2021dialogue} uses a language model with schema-driven prompting and proves that natural language prompt can effectively constrain model prediction, but this constraint is still not strong enough to prevent this situation from happening.  

% 数据集都是干什么的
% Most methods above are done using the general datasets like MultiWOZ(\cite{budzianowski2018large}, \cite{ramadan2018large}, \cite{eric2019multiwoz}, \cite{zang2020multiwoz}), Schema-Guided Dialogue Dataset \cite{rastogi2019towards}, etc. But few of these methods focus on the medical dialogue. 

\section{Task Definition}
% 图片 全抽象/全具象

A dialogue $D$ consists of a list of utterances  $U_1 $,  $R_1 $,..., $U_T$, $R_T$, where $U$ is the user's input and $R$ is the response of the system. $T$ is the total turns of the dialogue. The input of the t$_{th}$ turn is denoted as $D_t = \{U_1, R_1,..., R_{t-1}, U_t\}$. Notice that $R_t$ is not in $D_t$ because it will be the output of this turn for the system.

A \textit{schema} must be predefined to represent the structure of the state. The schema involves a set $I$ indicating the intents of the user, and a hierarchical structure $S$. An example can be found in Table \ref{state-sample}. $I$ contains all possible intents, like $\{describe, ask, ...\}$ for patient's intents and $\{diagnose, recommend, ...\}$ for doctor's intents. There are multiple hierarchies in $S$. The first hierarchy of $S$ contains all the fields $f$ needed for the system, like \textit{symptom}. The second hierarchy of $S$ is a set of the slot types required by the corresponding field, like the \textit{name} of the \textit{symptom}. The value of each field $V_f$ is a compound structure including all the slots in $S[f]$. 

% The first hierarchy is something like \textit{symptom}. The slot in the deeper hierarchy is used to describe the value of upper level instead of directly describing the state. For instance, the \textit{extent}, it's meaningless to directly put it into the state space, but it will be useful if we combine it with a \textit{symptom}.

% The first hierarchy of $S$ contains all the field $f$ needed for the system, and the second hierarchy of $S$ is the specific slot types required by the field. The value of each field $V_f$ is a compound structure including all the slots in $S[f]$.

% There has a function $F$ and for each $i$ in I, the result of $F(i)$ contains the related field to $i$. Using map $S$ and the field $f$ in $F(i)$, $S[f]$ is the target slot types set. The type of slot value $V_{s}$ can be one of the predefined result set it it's a categorial task or a span of the input if it's an extractive tasks. The value can also be a list if the target is a multiple-answer task. An example can be found in Table \ref{state-sample}.

Given a dialogue history ($D_t$) and a schema ($I$, $S$), the task is to first identify the intent $I_t$ of the input utterance $U_t$ and then identify all values $V_f$ for each field $f$ in $S$. The structure of $V_f$ is same with $S[f]$.
% The structure of $V_f$ is same with $S[f]$.

% If $V_{S_{I_t}} \ne \emptyset $ and $S_{S_i} \ne \emptyset$, the value of each slot in $S_{S_i}$ for each value in $V_{S_{I_t}}$ also needs to be extracted which form is similar with the slot value above.

\section{Approach}

\begin{figure*}[htbp] %H为当前位置，!htb为忽略美学标准，htbp为浮动图形
\centering %图片居中
\includegraphics[width=0.8\textwidth]{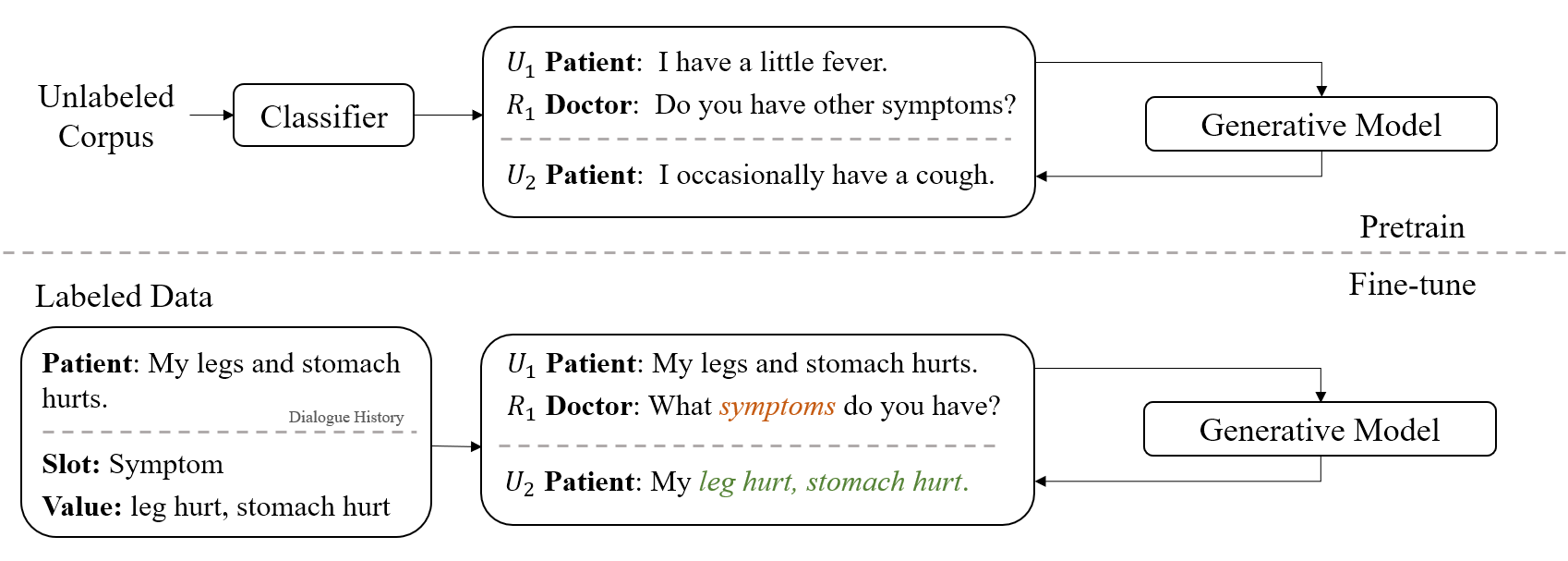} %插入图片，[]中设置图片大小，{}中是图片文件名
\caption{ Two-stage training process. The first stage is pretraining. The data for pretraining is filtered by the classifier. The training target is to generate the response according to the dialogue history. The second stage is fine-tuning which follows the same idea. We convert the slot and value into a pair of dialogue-style question and answer. The training target is still to generate the answer using the dialogue history including the constructed question. } %最终文档中希望显示的图片标题
\label{two-stage} %用于文内引用的标签
\end{figure*}

%And then the slot value can be easily extracted from the output. With the help of dialogue-style prompt, we unify the tasks of the two stage.
\begin{table*}
\centering
\begin{tabular}{llll}
\hline \textbf{Input Dialogue} & \textbf{Query Type} & \textbf{Constructed Query Sample} & \textbf{Output result} \\ \hline
\multirow{3}*{Patient: I have a bad headache.} & Type name & symptom & headache \\
 & Question & What symptoms does the patient have ? & headache \\
 & Dialogue-style Prompt & Doctor: What symptoms do you have ? & Patient: I have headache. \\
\hline
\end{tabular}
\caption{ Samples of different strategies to construct the query. }
\label{qa-sample}
\end{table*}

% \begin{figure}
%     \centering
%     \def\svgwidth{\columnwidth}
%     \input{image.pdf_tex}
%     \caption{Construction process for training data. The training data during the pretraining process is picked by a simple classifier to find the dialogue which last utterance is descriptive. The training data for fine-tuning is constructed from the labeled data.} 
% \end{figure}
% In this work, we formalize the DST task as a sequence to sequence task using a generative model. Given a dialogue history $D_i$, we convert both intent and slot types with their value into a series of question answer pair $(Q, A)$ in the form of dialogue style. The input of the model is $D_i$ concatenated with the $Q$ and the output is $A$. In this way, we transform the DST task into a response generation task which enables the model able to be pretrained from the unlabeled dialogue corpus.

% question $q$ and put $q$ along with $D_i$ into the model. We let the model generate the answer of that question and in this way we can get the target intent and slot value.

In this work, we treat the DST task as a Seq2seq task using a generative model. As shown in Figure \ref{arche}, we use prompt token $P$ to transform the target (intent or slot type) and their value into a series of dialogue-style question-answer pair $(Q, A)$. By using $Q$ to generate $A$, we convert the DST task into a response generation task. This makes it possible to pretrain the model with unlabeled dialogue datasets. The result of the DST task is acquired by parsing the generated output utterance. An extra decoding algorithm is used to guarantee the utterance is well-formed.

\subsection{Dialogue-style Prompt}
\label{ssec:prompt}
There have been two common ways to construct the QA pair. As shown in Table \ref{qa-sample}, the easiest way is to simply use the slot type name with its description(\cite{lee2021dialogue}). Another way is to construct a question (\cite{lin2021zero}, \cite{du-etal-2021-qa}, \cite{liu-etal-2020-event}). However, it is easy to find that the dialogue itself is in the form of question and answer. If we construct the QA pair in the medical dialogue form, namely the question looks like an inquiry from a physician, and the answer is an ordinary utterance from a patient, we may fully utilize the large unlabeled medical dialogue corpus. Based on the above intuition, we propose a dialogue-style Prompt, as shown in the last row of Table \ref{qa-sample}.

%That is to say that the constructed prompt is an utterance that contains a question about the target instead of just one question, and the output also has a specified pattern that is easy to be parsed. With the help of the generative model, we can treat the DST task as a response generation task which makes it possible to pretrain the model with a large unlabeled medical corpus.

For example, if the target is \textit{symptom} "headache" with \textit{extent} "bad", we can construct a series of question and answers. For the first turn, the question is "Doctor: what symptoms do you have", and let the model generate "Patient: I have a headache". The prefix indicates the role of the speaker, "headache" is the value of the target slot, and "I have" is also a prompt indicating the model to output the expected result.  For the next turn, we want to further get the \textit{extent} of the headache symptom. We put the former output value of the last turn namely "headache" into the question, "Doctor: How is your headache". The answer is "Patient: I feel bad". We choose to use multi-turn QA instead of one like Seq2seq-DU \cite{feng-etal-2021-sequence}, because we think that it would be difficult for the model to generate long sequences with a specified format accurately. 

% Constructing the question manually is the most common choice, but the template is varying and it's hard to decide which one is the best. Since we use the generative model as the backbone, it can not only generate answer but also generate the question. In this way, we can use the constructed question as the bootstrap and let the model choose the best prompt.

There also have been many ways to output the answer, like indicating the start and end position(\cite{du-etal-2021-qa}), tagging the input sequence (\cite{du-etal-2021-qa}), or generating the answer. We choose the generative model because it is better in tackling different kinds of output, no matter extractive or categorial. It would be hard for other models to be able to deal with both two tasks. Besides, the generated output must follow a predefined pattern to be easily parsed to a structure. For example, "I have a, b, c", The comma can only appear in the middle of the output as the separator.
%The output is still an utterance but has a pre-designed pattern like  "I have a, b, c". The simple form makes it easy to further extract the values from the generated utterance.

% As for the categorial slot, some methods(\cite{lin2021zero}) choose to put the candidates in the input and transform it as an extractive slot. Therefore, we choose to put extra constraints during the decoding progress to ensure the model will only output the valid result.  If the question has multiple answers, we put "," as the marker of separation between each answer during training, and split each answer using "," during inference. We believe that "," can work better than something else like a special token "[AND]" because it keeps the same form as the natural language which could benefit from the pre-trained language model. The model will output "none" if the target values are not mentioned or do not exist.

% With this design, the workflow of the model will be first identify intents, then query all slots related to the intent, and query the value of next hierarchy for each extracted slot values if that slot has next hierarchy. 

\subsection{Model}
\label{ssec:model}

We use UniLM \cite{dong2019unified} as the backbone of our approach. UniLM uses a special lower triangle mask that enables BERT to deal with the Seq2seq task. 

% The main reason is that most generative like GPT is unidirectional which means that it can only generate output according to the tokens before. In our approach, we want the model able to generate the question and also keep the dialogue order. This means that the output is between the dialogue and answer and a bidireactional model is vital. We further change the mask and let the lower triangle part to the middle. This makes the model able to generate question according to the both sides of input.

The input sequence follows the BERT-style: $[CLS] <D_t> [SEP] <A> [SEP]$, which $D$ is the concatenation of dialogue along with the constructed question $Q$, and [CLS] and [SEP] are special tokens used in BERT. Notice that we put a special token "[PATIENT]" and "[DOCTOR]" in front of each utterance to denote the speaker's role. During the training process, the complete $A$ is the input and a mask is used to control the computation of loss. And in the inference process, the model will generate the output in an autoregressive way which means the output of the last round will be the input of this round.

We use two-stage training as shown in Figure \ref{two-stage}. We first pretrain the model to make it adapt to medical dialogue and then fine-tune it on the downstream task with labeled data. With the help of the dialogue-style prompt, we unify the pretraining task and the downstream task by treating it as a response generation task.
 
 %The main idea of pretraining is to let the model able to generate the response according to the chat history so that it can also be able to answer the question constructed by the one mentioned in Section \ref{ssec:prompt}.
\textbf{Pretraining} There are different types of utterances in the unlabeled corpus. The patients may ask questions, describe their status, or acquire for comfort. We pick the utterances which are descriptive since the data are more similar to the downstream task. There is an intent type named "describe" in the annotated dataset. Using the dataset, we train a simple classifier to identify all descriptive utterances in the large unlabeled corpus. we also put some random utterances into the training data in order to make the model more adaptable. We mix the descriptive utterance and random one in a ratio of 4:1.

\textbf{Fine-Tune} During the fine-tuning process, we transform the labeled data into the dialogue style and let the model generate the constructed answer. 

\subsection{Decoding Strategy}

%The generate output must follow a predefined pattern to be easily parsed to a structure. For example, "," can only appear in the middle of the output as the separator. 

The major problem of the generative model is that the output is unpredictable and may be invalid. Furthermore, putting the history of dialogue into the input can help the model better understand the conversation but may introduce more noises \cite{yang-etal-2021-comprehensive} as well. Most generative-based DST methods simply discard invalid values. \cite{lin2021zero} uses canonicalization technique \cite{gao-etal-2020-machine} to replace the predicted value with the closest value in the ontology. However, ontology is not always available. 

% As mentioned in \ref{ssec:model}, we only focus on the latest turn of utterance, but we input the whole dialogue. This could also be achieved by only inputting the latest turn, but we want to give the model more context information, which may cause introducing noises.\cite{yang-etal-2021-comprehensive} shows that larger granularity can help generative decoding. 

% In most cases of generative model, there doesn't exist the limit of output. Like the answer generation and summary generation, generating any words will be reasonable. But in our study, generative model may generate nonsense result like words in old turns of dialogue or invalid value which is not in the categorical question's answer set. That's why greedy decoding algorithm and beam search are available for most generative task but not suitable in this work.

To prevent the situation from happening, we use a constrained decoding algorithm during inference. Since the model generates the result in an autoregressive way, it is easy to control the decoding progress. In each step of decoding, we only focus on a limited range of tokens instead of the whole vocabulary. In this way, we can ensure the output is in the expected pattern. 

For the extractive result, the output range will be limited to the tokens in the latest utterance, since the state change can only come from the newest input. For the categorical target, a trie-tree is constructed according to the candidates and we use it to guide the output range.

% Using this strategy, the influence caused by other utterances can be naturally eliminated.

%  This still cannot promise that the output is always valid and if the number of candidates is large, it will occupy the input space and as a result, lose the dialogue information. We construct a trie-tree according to the candidates and follow it to construct the output range. "," will be put into the output range if that's a multi-answer question, and "[SEP]" will be put into that when it's valid to end. By this means, a strong constraint is put on the output without modifying the input form.

% Notice that we only use this decoding algorithm during the inference time in order to avoid outputting invalid result, which is unnecessary when training. This could also be useful when transferring to a new domain, which stands a good chance to generate result in old domains which is invalid if no other methods are taken.

% \begin{table}
% \centering
% \begin{tabular}{lr}
% \hline 
% Dialogue Count & 89\\
% Total Utterance Count & 1478\\
% Avg. Tokens per Utterance & 26.72\\
% % Total Slot Count & 1151 \\
% \hline
% Patient Intent Category Count & 11\\
% Doctor Intent Category Count & 6\\
% Slot Category Count & 7\\
% \hline
% Total slot values & 1708 \\ 
% Discontinuous value & 165 \\
% Nested values & 26 \\
% \hline
% \end{tabular}
% \caption{\label{font-table} Statistics on the Medical DST dataset }
% \end{table}
\begin{table}
\centering
\begin{tabular}{lrrr}
\hline
 & MDST & SGD & MWOZ 2.2\\
\hline 
No. of domains & 1 & 16 & 8 \\
No. of dialogues & 89 & 16,142 & 8,438 \\ 
No. of utterances & 1478 & 329,964 & 113,556\\
Avg. tokens per turn & 26.72 & 20.44 & 13.46\\
% Total Slot Count & 1151 \\
\hline
Avg. No. of intents   $^{\triangle}$   & 11+6  $^{\diamondsuit}$ & 2.5 & 1.375 \\ 
Avg. No. of slots $^{\triangle}$ & 12 & 13.43 & 7.6 \\
No. of hierarchial slots & 5 & 0 & 0 \\
% \hline
% Total slot values & 1708 \\ 
% Discontinuous value & 165 \\
% Nested values & 26 \\
\hline
Discontinuous value & Yes & No & No \\
Slot with multi values & Yes & No & No \\
\hline
\end{tabular}
\caption{\label{font-table} Statistics on the Medical DST dataset comparing with Schema-Guided Dialogue Dataset \protect\cite{rastogi2019towards} and MultiWOZ 2.2 \protect\cite{zang2020multiwoz}. Our dataset MDST defines more fine-grained intents and hierarchical slots with discontinuous values.  $^{\triangle}$ We calculate the average number per domain for other datasets. $^{\diamondsuit}$ 11 for patient intents and 6 for doctor intents. }
\end{table}
% 和别的比较 能显出难点的地方
% 分布不一样 和 challenge 呼应

\section{Experiments}

\subsection{Dataset}

% Medical dialogue state tracking mainly has two different challenges: i) the state has multiple hierarchies, ii) the slot value might be discontinuous in the utterance. To the best of our knowledge, no public datasets consider this kind of situation. Therefore, we annotate and publish a Medical Dialogue State Tracking dataset. We also use another medical slot filling dataset which is a subset of our task to evaluate our approach.

% \textbf{MDST} is the dataset annotated by us. The intent and slot are defined with the decision support part and with the help of clinicians. All the dialogues were the record of conversation published by the online consultation service provider. All response received by the patients are written by professional doctors. We randomly annotate part of them and others are used for pretraining. Consecutive utterances from the same people are concatenated into one. Detailed statistics are shown in Table \ref{font-table}.

\textbf{MDST} is the dataset annotated in this paper. The original corpus comes from one of the biggest online medical consultation service providers in China, who hired clinicians to give suggestions to patients on the Internet. We collect 183,386 dialogues and annotate 89 of them. Detailed statistics are shown in Table \ref{font-table}. Compared to the two commonly used DST datasets, the dataset is much smaller. However, it costs a lot of effort to define the annotation specification with the help of clinicians. It also takes about six persons more than 2 months work to make the annotation applicable. The dataset is divided into training, test, and validation sets in a ratio of 8:1:1.

%Consecutive utterances from the same people are concatenated into one.
% The dataset involves intents and slots two parts. One utterance might have multiple intents like "I have a headache. What medicine should I take?" has \textit{describe} and \textit{ask} two intents. Each intent has its own related slots, despite the slots of different intents might overlap. The slot value could be categorical like \textit{yes} and \textit{no} or extractive like part of the utterance. The slot value can be plain or be combined into a list or a structure with multiple hierarchies. Detail statistics are show in Table \ref{font-table}. The dataset is divided in a ratio of 8:1:1.

\textbf{MSL} is proposed by \cite{Shi_Hu_Che_Sun_Liu_Huang_2020} which is a dataset for medical slot filling task. It only focuses on \textit{symptom} slot with finite values. It contains 1152 labeled utterances for training, 500 for validation, and 1000 for the test. We keep the same setup in our experiments.

\begin{table}
\centering
\begin{tabular}{lrrr}
\hline \textbf{Model} & \textbf{Precision} & \textbf{Recall} & \textbf{F1} \\ \hline
BERT & 76.96 & 64.92 & 70.44 \\ 
QA w. typename & 80.40 & 75.83 & 78.05 \\
QA w. question & \textbf{80.48} & \textbf{80.09} & \textbf{80.29} \\
\hline
Ours & 79.29 & 74.41 & 76.77 \\
\quad + Pretrain & 80.30 & 75.36 & 77.75 \\
\hline
\hline
Ours$_{80\%}$ & 79.14 & 70.14 & 74.37 \\
\quad + Pretrain$_{80\%}$ & 78.33 & 75.36 & \textbf{76.81} \\ 
\hline
Ours$_{50\%}$ & 78.65 & 71.56 & 74.94 \\
\quad + Pretrain$_{50\%}$ & 78.87 & 72.51 & \textbf{75.56}  \\
\hline
Ours$_{20\%}$ & 60.70 & 57.82 & 59.22 \\
\quad + Pretrain$_{20\%}$ & 64.29 & 59.72 & \textbf{61.92} \\
\hline
\end{tabular}
\caption{\label{intent} \label{intent-few}  Intent Detection Experiment Results. The first two rows of the table show the performance of our methods compared with the baselines. The last three rows show the performance with different ratios of training data.}
\end{table}

\begin{table}
\centering
\begin{tabular}{lrrrr}
\hline \textbf{Model} & \textbf{Precision} & \textbf{Recall} & \textbf{F1}\\ \hline
QA w. type name & 46.30 & 28.82 & 35.52 \\
QA w. question& 43.64 & 29.68 & 35.33 \\
\hline
Ours & 49.41 & \textbf{36.02} & 41.67 \\
\quad + Pretrain & \textbf{51.03} & 35.73 & \textbf{42.03} \\ 
\hline
\hline
Ours$_{80\%}$ & 48.89 & 31.70 & 38.46 \\
\quad + Pretrain$_{80\%}$ & 46.01 & 34.87 & \textbf{39.67} \\ 
\hline
Ours$_{50\%}$ & 45.54 & 29.39 & 35.73 \\
\quad + Pretrain$_{50\%}$ & 49.77 & 31.12 & \textbf{38.30}  \\
\hline

Ours$_{20\%}$ & 42.27 & 26.80 & 32.80 \\
\quad + Pretrain$_{20\%}$ & 46.52 & 30.84 & \textbf{37.09} \\
\hline
\end{tabular}
\caption{\label{slot} \label{slot-few} Slot Filling Experiment Results. The first two rows of the table show that our method outperforms the baselines. The last three rows show the performance with different ratios of training data. Pretraining can provide greater improvement to the model in the scenario with less data. }
\end{table}

\subsection{Baseline}
We set baselines of the intent detection and the slot filling tasks separately since few researchers process the two tasks in one unified model as we are. 
    
\textbf{BERT} \cite{devlin-etal-2019-bert} is easy to fine-tune on the intent detection task if we treat the task as a multi-label classification task. The input is one utterance instead of the whole dialogue history and the output is the intent of that utterance.

\textbf{QA} We follow the method proposed by \cite{lin2021zero} as the baseline. It treats the slot filling task as a Seq2Seq task in the form of question answering. We use the two methods in Table \ref{qa-sample}. One is to simply use the target slot type name, and the other is to convert the type name into a question with a template.

\begin{table*}
\centering
\begin{tabular}{lrrrrr}
\hline \textbf{Model}& \textbf{Precision}& \textbf{Recall} & \textbf{Micro F1} & \textbf{Macro F1} & \textbf{Turn Accuracy}\\ 
\hline
DRNN+A+WS & 82.94 & 79.44 & 81.15 & 76.95 & 58.30 \\
TextCNN+BERT+WS & - & - & 75.56 & 57.17 & 52.80 \\
\hline
Ours & 88.93 & 87.80 & \textbf{88.36} & 86.83 & 78.90\\
% Prompt$_{cate}$ & 88.66 & 87.80 & 88.23 & 86.66 & 78.80\\
% Prompt$_{extr}$ & 91.56 & 58.88 & 71.67 & 73.22 & 50.60 \\
% % Prompt_{beam no throw} & 89.04 & 88.97 & 89.00 & 73.22 & 77.40 \\
% Prompt$_{beam no throw}$ & - & - & - & - & 77.40 \\
% \hline
\quad + Pretrain & 89.11 & 87.57 & 88.34 & \textbf{87.75} & \textbf{79.60} \\
% w. Decoding$_{extr}$ & 92.57 & 59.11 & 72.15 & 73.92 & 51.50 \\
% w. Decoding$_{cate}$ & 88.73 & 87.80 & 88.26 &87.58 & 79.30\\
% % w. Decoding_{beam no throw} & 89.06 & 88.11 & 88.59 &87.96 & 78.20\\
% w. Decoding$_{beam no throw}$ & - & - & - & - & 78.20\\
\hline
\end{tabular}
\caption{\label{msl} Results on MSL. 'A' respresents lable-attentive model and ‘WS’ represents weak supervision methods which are from \protect\cite{Shi_Hu_Che_Sun_Liu_Huang_2020} along with the performance results. }
\end{table*}

\begin{table}
\centering
\begin{tabular}{lrrrr}
\hline \textbf{Model} & \textbf{Precision} & \textbf{Recall} & \textbf{F1}\\ \hline
BERT+CRF & 63.41 & \textbf{61.17} & \textbf{62.27} \\
Ours & \textbf{63.49} & 53.33 & 57.97 \\
\quad + Pretrain & 61.76 & 54.69 & 58.01 \\

\hline
\end{tabular}
\caption{\label{ner-perf} Performance on simplified entity recognition task.}
\end{table}

\subsection{Metrics}

\textbf{Intent Detection} could be defined as a multi-label classification task, so we use precision, recall, and f1 as the metrics. 

% \textbf{Slot Filling} is a bit complicated. We first transform the multi-hierarchies state into a flat structure like a tuple list which style just like ("symptom", "headache") and then calculate the precision, recall, and f1 between the predicted tuple list and answer tuple list. As for the slots with multiple values, they will be transformed into multiple tuples. And for the slots with hierarchies, they will be flattened with the central word like ("symptom", "headache", "extent", "serious"). With the approach, only when the slot value is completely the same as the answer is seen as correct which means we evaluate the answer in the word level instead of token level like the PRF evaluation metrics in most NER tasks.

\textbf{Slot Filling} is a bit complicated. We first transform the hierarchical state into a flat structure like ("symptom", "headache") and ("symptom", "headache", "extent", "serious"). In this way, it would be convenient to calculate the precision, recall, and f1 between the predicted tuple list and answer tuple list. With the approach, only when the slot value is completely the same as the answer is regarded as correct. 

% We keep the same evaluation method with the setup of other datasets.

\subsection{Setup}

To ensure the comparison is fair, our approach and all baseline models use BERT-base and \textit{chinese\_roberta\_www\_ext} from \cite{DBLP:journals/corr/abs-1906-08101} as the pre-trained weights. We use cross-entropy as the loss function and Adam as the optimizer with the learning rate of 1e-5. The batch size is 16 and each epoch has 1000 steps, 30 epochs in total. All the experiments are done on one NVIDIA Geforce RTX 3090 with 24GB VRAM.

\subsection{Experiment Results}

According to the first two rows of Table \ref{intent}, we can see the generation-based methods outperform the simple classification model. This shows that generation-based methods are suitable for the intent detection task. However, the performance of our approach is 2.54\% worse than QA, only 0.18\% on precision but 4.73\% on recall. But from the first two rows of Table \ref{slot}, our approach is 6.51\% superior to the baseline on slot filling task. This result proves that our approach is more suitable for the more complicated task with larger state spaces.

From Table \ref{msl} we can see that our method is much better than the result reported by \cite{Shi_Hu_Che_Sun_Liu_Huang_2020} in the MSL dataset. Although it's possible to treat this slot filling dataset as a multi-label classification problem like \cite{Shi_Hu_Che_Sun_Liu_Huang_2020}, our approach is still suitable for this task, and pretraining can almost always improve the performance.

% \begin{table*}
% \centering
% \begin{tabular}{p{1cm}p{4cm}p{5cm}p{5cm}}

% \hline \textbf{ID} & \textbf{Utterance} & \textbf{Predicted Result} & \textbf{Annotated Result}\\ \hline
% 1 & 
% \begin{CJK*}{UTF8}{gbsn} 一般是手痛，也有其他地方\end{CJK*} \newline Hand usually hurts, but also elsewhere & 
% \begin{CJK*}{UTF8}{gbsn} ('Symptom', '手痛 (hand hurts)') \newline ('手痛 (hand hurts)', 'Body-part', '手(Hand)') \end{CJK*} &
% \begin{CJK*}{UTF8}{gbsn} ('Symptom', '手痛 (hand hurts)') \newline ('手痛 (hand hurts)', 'Body-part', '手(Hand)') \end{CJK*} \\
% \hline
% 2 & 
% \begin{CJK*}{UTF8}{gbsn} 还有鼻塞，但是站起来不塞\end{CJK*} \newline And a stuffy nose, but not when standing up & 
% \begin{CJK*}{UTF8}{gbsn} ('Symptom', '鼻塞 (stuffy nose)') \newline ('鼻塞 (stuffy nose)', 'Body-part', '鼻(nose)') \end{CJK*} &
% \begin{CJK*}{UTF8}{gbsn} ('Symptom', '鼻塞 (stuffy nose)') \end{CJK*} \\
% \hline

% \hline

% \end{tabular}

% \caption{\label{err} Error examples }
% \end{table*}

% \begin{figure}[htbp] %H为当前位置，!htb为忽略美学标准，htbp为浮动图形
% \centering %图片居中
% \includegraphics[width=0.4\textwidth]{loss.png} %插入图片，[]中设置图片大小，{}中是图片文件名
% \caption{Record of training loss. The experiments follow the same setup except that whether using pretrained weights to initialize.} %最终文档中希望显示的图片标题
% \label{loss} %用于文内引用的标签
% \end{figure}

\subsection{Ablation Study}

\textbf{Dialogue-style Prompt} QA method uses pairs of questions and answers without the dialogue-style prompt. From Table \ref{intent} we can find that the performance of our approach is 2.54\% worse than QA on intent detection task. But in the slot filling task, our approach outperforms 6.51\% to the QA method. This may suggest that the dialogue-style prompt could make the model focus on the semantic information so that it can help improve the performance on complex slot filling tasks. For the simple task like intent detection, the help is minimal. Another possible reason is that the definition of intent does not completely conform to the normal speech habits which makes the prompt confuse the model.

\textbf{Pretraining} To find out whether the pretraining can help the model understand the dialogue, we trained the model with the same setup but loaded the parameters of BERT instead of the pre-trained weights. From Table \ref{intent} and \ref{slot}, we can see that pretraining has slight help in intent detection but large help in slot filling. The result is consistent with the expectation that pretraining is able to make the model find out the relationship between the utterances. This is more helpful for the difficult task. The result on the MSL dataset in Table \ref{msl} also supports our conclusion.

We find that loss decreases faster during the fine-tuning process with pretraining. One of the main reasons is that the pretraining has made the model learn to generate utterances while the original BERT cannot. This implies pretraining might be helpful in low-resource scenarios which are common in the medical field.

To prove the conjecture above, we reduce the training dataset to 20\%, 50\%, and 80\% and repeat the training process. From Table \ref{intent-few} and \ref{slot-few}, we can see that the size of training data has a great impact on the performance, and the model with pretraining performs better than the one without pretraining in most scenarios. This proves that our pretraining approach is helpful in low-resource scenarios. For the simple task like intent detection, when the data is sufficient, the impact of pretraining becomes smaller like the result between 80\% and 100\% data in Table \ref{intent-few}.

\subsection{Error Analysis}

We further look into the error produced by the model and compare the result with the BERT+CRF which is a common but effective NER model. Notice that since NER cannot handle discontinuous and multi-hierarchies structures, we only compare the performance on the first-level slots. According to Table \ref{ner-perf}, we notice that the performance of our approach is a bit lower than BERT+CRF and the main gap is the recall rate.

%We further look into the erresult with the BERT+CRF which is a common but effective NER model. Notice that since NER cannot handle discontinuous and multi-hierarchies structures, we only compare the performance on the first-level slots. According to Table \ref{ner-perf}, we notice that the performance of our approach is a bit lower than BERT+CRF and the main gap is the recall rate.

\begin{CJK*}{UTF8}{gbsn}
We find two typical scenarios to explain the reason. For example, the output of the utterance 我肚子感觉难受 (My stomach feels uncomfortable) is (\textit{symptom}, 肚子感觉难受 "stomach feels uncomfortable") which is correct for the NER model. But our approach outputs (\textit{symptom}, 肚子难受 ("stomach uncomfortable") which extracts a discontinuous entity. The answer is also correct. However, it is hard for us to find all possible correct answers and put them in the annotated corpus. This example may partly explain why the recall performance is not that ideal. Another example is the \textit{symptom} 皮 肤 过 敏(skin allergy) whose \textit{body\_part} is 皮 肤(skin). 
Our approach extracts 过 敏(allergy) as a \textit{symptom} whose \textit{body\_part} is 皮 肤(skin). It is also hard to decide whether the two should be split.  
\end{CJK*}

\section{Conclusion}

In this work, we first redefine the state with multiple hierarchies and annotate a dataset called MDST extending existing DST tasks. We propose a dialogue-style prompt with UniLM to solve the new problem. The proposed prompt also makes it possible to pretrain the model with the unlabeled medical dialogue corpus. The experiments show that our method outperforms baseline up to 6.51\%, and in low-resource scenarios, the pretraining can improve the performance up to 4.29\%.

% In this work, we propose that the existing DST method cannot fully handle the problems in medical dialogue systems, specifically in the following three parts: i) the state definition of most current DST task is not fine-grained enough, ii) the colloquial input makes the slot values scattered, iii) learning-based methods require a large annotated task which is not available in the medical field.  Therefore, we first redefine the state with multiple hierarchies and annotate a dataset. We propose to use the dialogue-style prompt with a generative model to solve the discontinuous values. The proposed prompt also makes it possible to pretrain the model with the medical dialogue corpus. The experiments show that our method outperforms other methods, especially in low-resource scenarios.

\bibliographystyle{named}
\bibliography{ijcai22}

\end{document}